\documentclass{article}

\usepackage{arxiv}

\usepackage[utf8]{inputenc} 
\usepackage[T1]{fontenc}    
\usepackage{hyperref}       
\usepackage{url}            
\usepackage{booktabs}       
\usepackage{amsfonts}       
\usepackage{nicefrac}       
\usepackage{microtype}      
\usepackage{lipsum}		
\usepackage{graphicx}
\usepackage{natbib}
\usepackage{doi}
\usepackage{tabularx}
\usepackage{multirow}
\usepackage{amsmath}
\usepackage{amssymb}

\title{A systematic study of the foreground-background imbalance problem in deep learning for object detection}

\author{{Hanxue Gu}\\
	Electrical and Engineering Department\\
	Duke University\\
	Durham, NC, 27703 \\
	\texttt{hanxue.gu@duke.edu} \\
	\And
	{Haoyu Dong} \\
	Electrical and Engineering Department\\
	Duke University\\
	Durham, NC, 27703 \\
	\texttt{haoyu.dong@duke.edu} \\
     \And
 	{Nicholas Konz}\\
	Electrical and Engineering Department\\
	Duke University\\
	Durham, NC, 27703 \\
	\texttt{nicholas.konz@duke.edu} \\
 \And 
  	{Maciej A. Mazurowski}\thanks{Maciej A. Mazurowski also has secondary appointments in the Department of Computer Science; Department of Electrical and Engineering Department; Department of Biostatistics and Bioinformatics at Duke University.} \\
	Department of Radiology\\
	Duke University\\
	Durham, NC, 27703 \\
	\texttt{maciej.mazurowski@duke.edu} \\
}

\renewcommand{\shorttitle}{\textit{arXiv} Template}

\hypersetup{
pdftitle={A template for the arxiv style},
pdfsubject={q-bio.NC, q-bio.QM},
pdfauthor={David S.~Hippocampus, Elias D.~Striatum},
pdfkeywords={First keyword, Second keyword, More},
}

\begin{document}
\maketitle

\begin{abstract}
The class imbalance problem in deep learning has been explored in several studies, but there has yet to be a systematic analysis of this phenomenon in object detection. Here, we present comprehensive analyses and experiments of the foreground-background (F-B) imbalance problem in object detection, which is very common and caused by small, infrequent objects of interest. We experimentally study the effects of different aspects of F-B imbalance (object size, number of objects, dataset size, object type) on detection performance. In addition, we also compare 9 leading methods for addressing this problem, including Faster-RCNN, SSD, OHEM, Libra-RCNN, Focal-Loss, GHM, PISA, YOLO-v3, and GFL with a range of datasets from different imaging domains. We conclude that (1) the F-B imbalance can indeed cause a significant drop in detection performance, (2) The detection performance is more affected by F-B imbalance when fewer training data are available, (3) in most cases, decreasing object size leads to larger performance drop than decreasing number of objects, given the same change in the ratio of object pixels to non-object pixels, (6) among all selected methods, Libra-RCNN and PISA demonstrate the best performance in addressing the issue of F-B imbalance.  (7) When the training dataset size is large, the choice of method is not impactful (8) Soft-sampling methods, including focal-loss, GHM, and GFL, perform fairly well on average but are relatively unstable.
\end{abstract}


\section{Introduction}

Object detection (OD) is a well-studied and widely used computer vision task. Examples include localization of traffic signs, vehicles, and pedestrians in autonomous driving, screening damaged and aberrant equipment, and identifying abnormalities in medical imaging. Many methods have been proposed over time for this task, with deep learning now being the most popular methodology, showing impressive performance across different applications.
Several challenges impede the application of deep learning to the task of object detection, including the fact that (1) the target objects are often small in relation to the image size and (2) the number of objects present in the training data may be small. We refer to these problems jointly as foreground-background (F-B) class imbalance to reflect the common scenario when the number of pixels that belong to target objects is smaller than those belonging to the background. A related class imbalance problem caused by a limited number of examples of certain classes compared to others has been extensively studied in the literature on classification tasks \cite{liu2008exploratory,van2007experimental,huang2016learning,buda2018systematic}. Still, the problem caused by a limited number of foreground pixels in object detection scenarios has received less attention. While some approaches have been proposed to address the inter-class imbalance between the number of candidates from foreground and background classes or between foreground classes in the context of object detection \cite{qian2020dr,shrivastava2016training,pang2019libra,cao2020prime}, the studies are limited to showing performance improvements and do not provide a thorough understanding of the impact of the imbalance and its different components. The same is the case for previous attempts at literature reviews on this topic \cite{chen2020foreground}, which are restricted to a literature study listing the techniques and do not  provide an adequate analysis of the causes of the F-B imbalance and consequences on model performance. Other relevant works are \cite{liu2021survey,tong2020recent,nguyen2020evaluation,liu2020survey}, which focus on small object detection. Although small objects can cause the F-B imbalance problem in certain scenarios, other factors, such as the limited number of objects or sparse objectives, can also lead to the same problem. Moreover, these studies are confined to a methodological categorization. Most of them contain no experimental comparison, making it difficult to identify the most effective strategy when confronting the problem of F-B imbalance. For the others, they \cite{liu2020survey,tong2020recent,oksuz2020imbalance} only present experimental results published in original work on commonly used object detection datasets, such as COCO \cite{lin2014microsoft} and PASCAL \cite{everingham2010pascal}. Due to the complexity of these datasets, which include a large number of item categories and various object sizes (seen in Figure \ref{fig:coco-fig}), and where a large proportion of objects are actually under relatively large object sizes and dense object distributions, it is virtually impossible to disentangle the extent to which these techniques resolve the problem of F-B imbalance and the methodologies' overall performance of addressing imbalance issue on these datasets. To the best of our knowledge, no previous studies provide a comprehensive understanding of the impact of F-B imbalance and its various sources and a comprehensive comparison of potential methods in the context of F-B imbalance and provide a taxonomy for these methods.

\begin{figure*}
    \centering
    \includegraphics[width=\columnwidth]{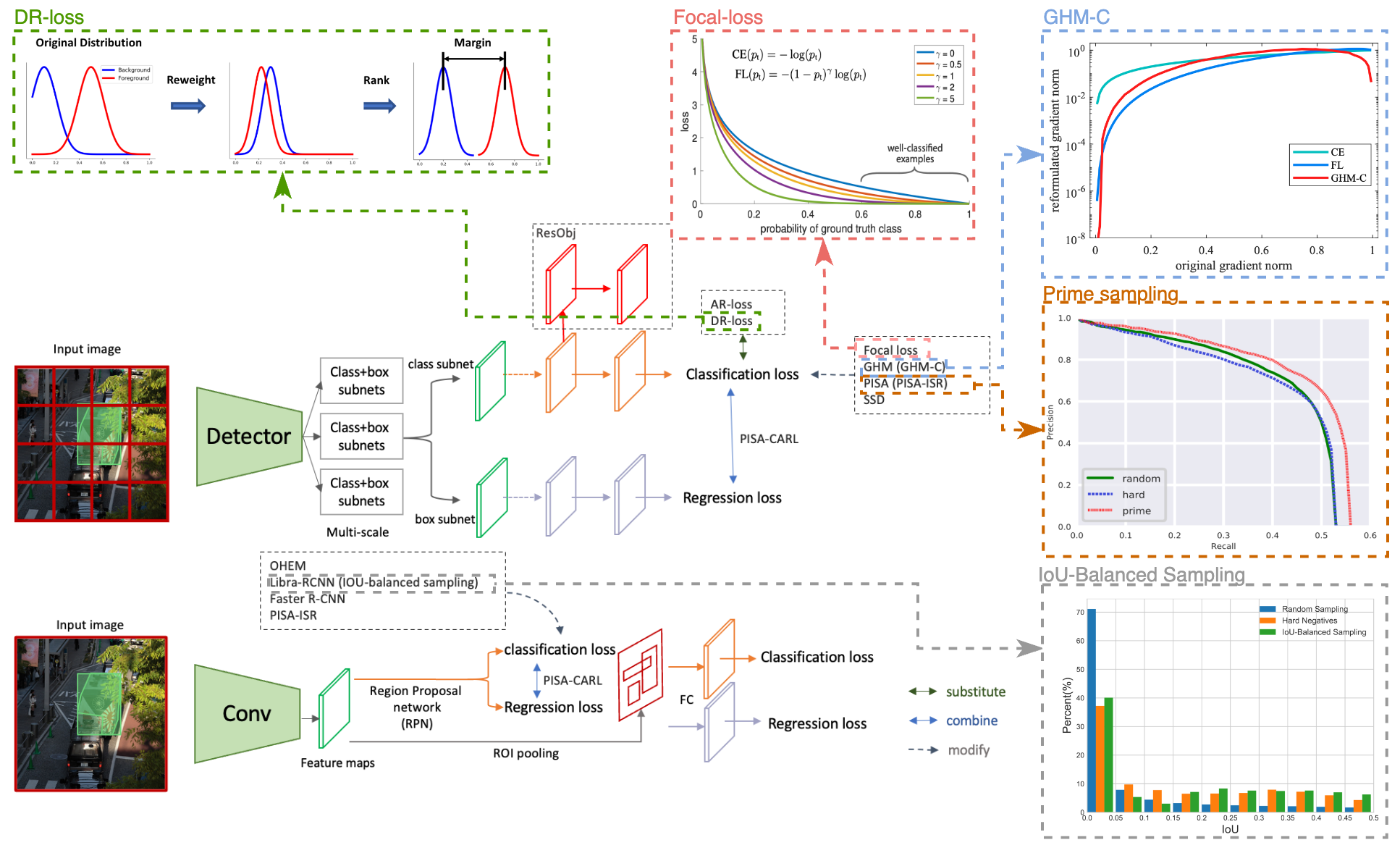}
    \caption{An illustration of 1-stage (above) and 2-stage (below) detectors and the scope of action of different methods for dealing with F-B imbalance.}
    \label{fig:illustrate}
\end{figure*}

\begin{figure}
    \centering
    \includegraphics[width=1\columnwidth]{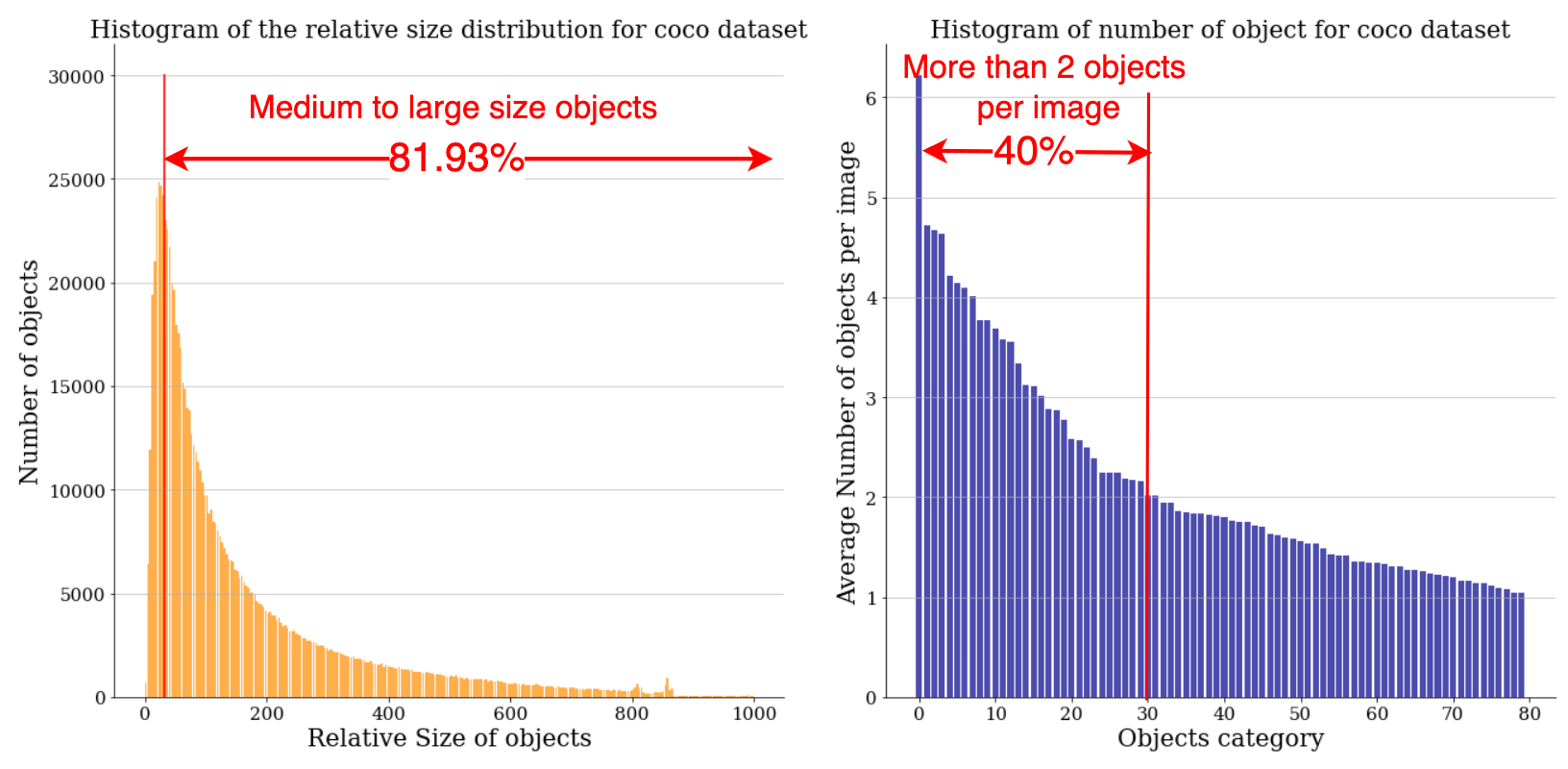}
    \caption{A visualization of the size distributions of the objects in the COCO dataset, where object size is defined as the average of width ($W_{obj}$) and height ($H_{obj}$), $S=\sqrt{W_{obj}*H_{obj}}$. (Left) is the distribution of relative object sizes, which is computed after resizing image sizes to a maximum of 1000 pixels in height or width. (Right) is the number of objects per image. Based on the small, medium, and large object definitions on the COCO challenge, 81.93\% of the objects in the COCO dataset are medium to large objects; and 40\% of the object categories have over 2 objects per image. The F-B imbalance varies a lot for different object categories owing to the various object sizes and object numbers.}
    \label{fig:coco-fig}
\end{figure}

\begin{enumerate}
    \item In our work, we give a thorough and exhaustive analysis of the F-B imbalance problem and its solutions in deep learning-based detection, supported by a large number of carefully crafted experiments. Specifically: We define the foreground and background (F-B) imbalance issues in object detection and identify the influencing factors. Importantly, we propose a definition of F-B imbalance that reflects the dataset's characteristics and is independent of the choice of detection methods.
    \item We experimentally evaluate the factors influencing F-B imbalance by creating synthetic datasets that control multiple variables and designing many experiments (over 4000 training sessions). To the best of our knowledge, this is the first to analyze the effects of this scale's imbalance problem. 
    \item We present a literature review on existing works to mitigate the F-B imbalance problem and construct a road map that classifies these strategies in a systematic manner, as shown in Figure \ref{fig:illustrate}.
    \item We implement 9 representative object detection algorithms and evaluate their performance with different metrics and data sets to provide a holistic guide for detection model selection.
    \item We summarize our findings with 13 statements regarding dataset imbalance and method selection. These findings were tested on the synthesis datasets, two popular natural object detection datasets, and a challenging medical imaging dataset and have broad applicability.
\end{enumerate}

\section{Definitions}

\begin{figure*}
    \centering
    \includegraphics[width=1\columnwidth]{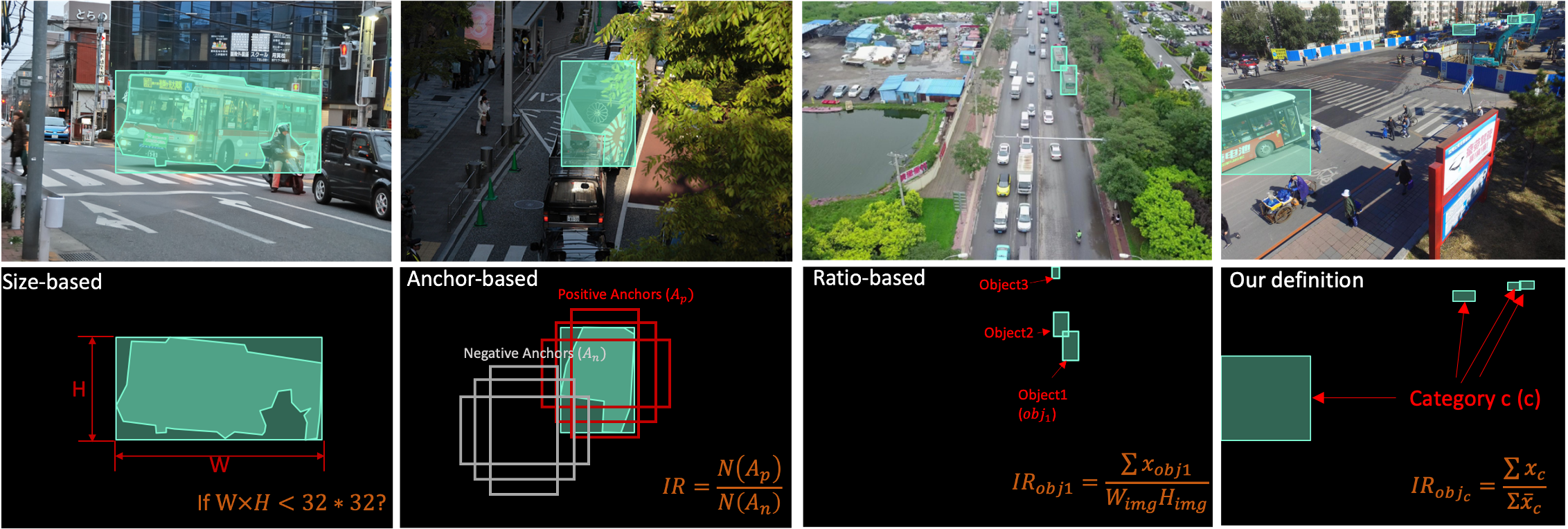}
    \caption{Illustration of the size and number of object distribution discrepancies among various datasets and F-B imbalance definitions. In the first row, the images on the left and center are from COCO, while the image on the right is from Vis-Drone, illustrating the variation in size and number of the same object category (bus) between images and datasets. The second row displays the ground truth bounding boxes corresponding to each image and exhibits several ways to define the F-B imbalance.}
    \label{fig:definition}
\end{figure*}

The general perception of foreground and background imbalance is that the target objects (foreground) usually only occupy a small region of the images compared to the areas not occupied by the target objects (background). 
Many studies have attempted to alleviate this problem, but there is no clear and unified definition of the F-B imbalance. For example, some studies focus on small object sizes, while others focus on the limited number of anchors. 

We categorize these conceptualizations that partially capture the problem of F-B imbalance in object detection into \textit{Size-based} definition, \textit{Anchor-based} definition, and \textit{Ratio-based} definition, and then propose a simple yet more comprehensive definition of imbalance that we believe will allow for a better understanding of the effects of class imbalance on detection performance in deep learning.

\subsection{Size-based imbalance – small objects}
\cite{liu2021survey,nguyen2020evaluation} mainly focus on the foreground and background imbalance introduced by the small objects: the small object detection problem, where the definition of imbalance is restricted to the size of a given object. Note that the term “small” here refers to the relative size compared to the size of the image rather than the actual size of the objects. \cite{torralba200880} supposed small objects are less than 32x32 pixels. This convention has been used by many popular dataset challenges, i.e., COCO \cite{lin2014microsoft}, the more challenging small objects are defined through object size in relation to the image size. While object size is crucial for object detection, focusing solely on it neglects the matter of the number of objects in the image. Even though each individual item is small, the population of objects might nevertheless fill a substantial percentage of the image.

\subsection{Ratio-based imbalance}
A similar convention is to use the ratio of the size of bounding boxes relative to the image size to define the F-B imbalance ratio ($IR$). For example, \cite{zhu2016traffic} identified the objects whose bounding box comprises less than 20\% of the entire image as F-B imbalanced objects and \cite{chen2017r} specified low mean relative overlap (i.e., the region covered by one object comprises a small proportion of the entire image) as F-B imbalance, 
\begin{equation}
    IR_{obj_i}=\frac{\Sigma x_{obj_i}}{W_{img} H_{img}}
\end{equation}
where $W_{img}$ and $H_{img}$ are the object width and height and $\Sigma x_{obj}$  is the aggregation of pixels belong to this single object $obj_i$, see Figure 1 (right). In \cite{chen2017r}, they specified that if $IR_{obj_i}$ is less than 0.58\%, this object faces an F-B imbalance and challenges the detection.

This definition is conventional, yet it disregards the potential consequences of the number of same-category items in an image since it focuses on each individual object rather than a certain type of object as well. Considering it from an imbalance perspective, it is worth pondering whether the F-B imbalance still exists if there are small items with a high index density (plenty of samples per image). Since the ratio-based definition has been widely embraced and applied in recent years, we adapted and changed it slightly in our work, defining imbalance not for a particular object item but as a characteristic of a whole dataset.

\subsection{Anchor-based imbalance}
\begin{figure*}
    \centering
    \includegraphics[width=1\columnwidth]{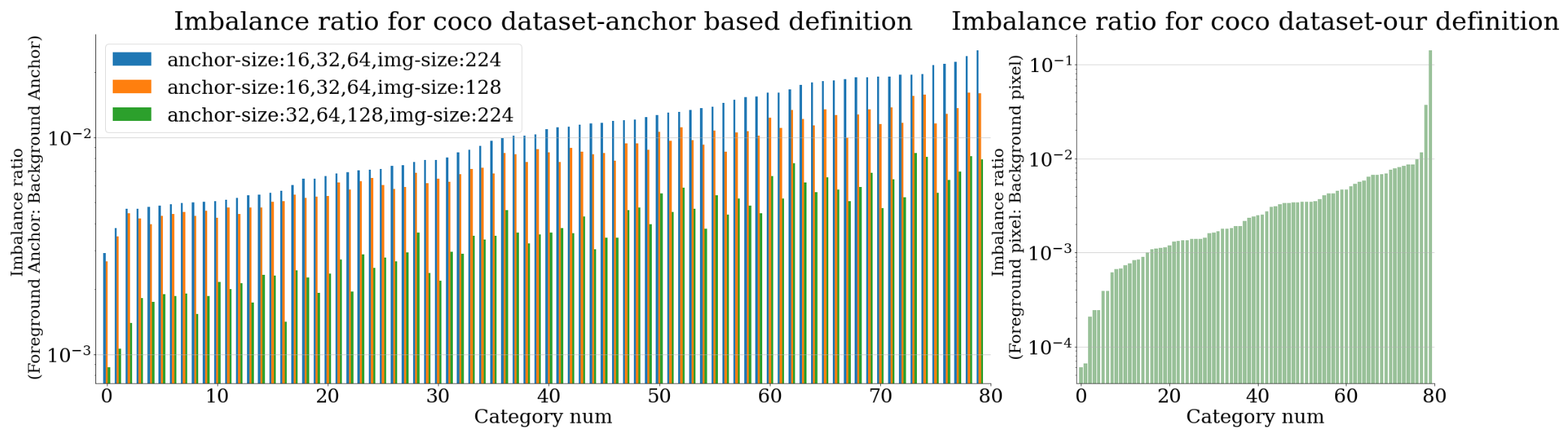}
    \caption{F-B imbalance ratio for COCO dataset under (1) the anchor-based definition (section xxx) on a. anchor size as 16, 32, 64 and image size as 224; b. anchor sizes as 16, 32, 64 and image size as 128; c. anchor sizes as 32, 64, 128 and image size as 224. (2) our definition, seen in section xxx (categories sorted by the descending imbalance ratio).}
    \label{fig:IR}
\end{figure*}

Most object detection methods introduce anchors \cite{ren2015faster}, which are a set of pre-defined boxes with a given width and height tiled across the images as the initial box locations. If an anchor's intersection over union (IoU) with the ground truth box is larger than a predetermined threshold, this anchor is usually considered to be in the foreground. Otherwise, it becomes an anchor for the background. Following this definition, \cite{He_2014,everingham2010pascal} formalized the F-B imbalance factor as the ratio of the number of foreground anchors to the number of background anchors in an anchor-based technique, 
\begin{equation}
    IR = \frac{N(A_p )}{N(A_n )}.
\end{equation}
As depicted in Figure \ref{fig:definition} (middle), $A_p$ and $A_n$ represent positive and negative anchors, respectively. This definition automatically limited its use to anchor-based detection approaches while excluding alternative anchor-free techniques. Figure \ref{fig:IR} (left) illustrates that the selection of pre-defined anchor sizes, the image size compared with the anchor size, and the threshold used to separate positives from negatives also alter this imbalance. It is preferable for the F-B imbalance ratio to be an inherent characteristic of the dataset that is unaffected by the selection of methods and hyperparameters.

\subsection{A new definition of F-B imbalance}
We propose a new, simple definition that comprehensively characterizes the F-B imbalance. Specifically, we define the imbalance between the foreground and background as the low ratio between the total number of pixels occupied by foreground objects and the number of pixels occupied by the background. 

To better parameterize the imbalance, we define the background pixel vs. object pixel imbalance ratio as
\begin{equation}
    IR_c = \frac{\Sigma x_c}{\Sigma \bar{x}_{c}}  \sim f(M_c,S_c),
\end{equation}
where $x_c$ are the pixels that belong to category $c$, whereas $\bar{x}_c$ are the pixels that do not belong to category $c$. We define that the smaller the $IR_c$, the more severe the F-B imbalance of category $c$.
Note that $IR_c \sim f(M_c,S_c)$ is influenced mainly by two parameters: the average ratio between the size of a single object and image ($S_c$) and the average number of objects per image ($M_c$) among the category $c$. The small object detection problem, which has been investigated extensively, arises if $S_c$ is small, and it makes $IR_c$ decrease. Additionally, $IR_c$ is reduced if there are only a few objects in each image. The imbalance ratio (IR) across categories under our definition for the coco dataset is represented in Figure \ref{fig:IR} (right). More than $50\%$ of the categories of objects have an $IR_c$ smaller than $10^{-2}$, demonstrating that F-B imbalance is quite common in natural scenarios. 

In the following sections, we refer to the $IR_c$ according to the F-B imbalance ratio, which we had defined here. Also, the object size we mentioned in this work is a relative ratio, which is calculated by $S_c = \sqrt{\frac{1}{N} \Sigma_{i=1}^N \mathbb{E}R_{c,i}}$, where N is the number of the objects belonging to category c and $R_{c,i}$ is the area belonging to object i.

\section{Methods for addressing F-B imbalance}
\begin{figure*}
    \centering
    \includegraphics[width=\columnwidth]{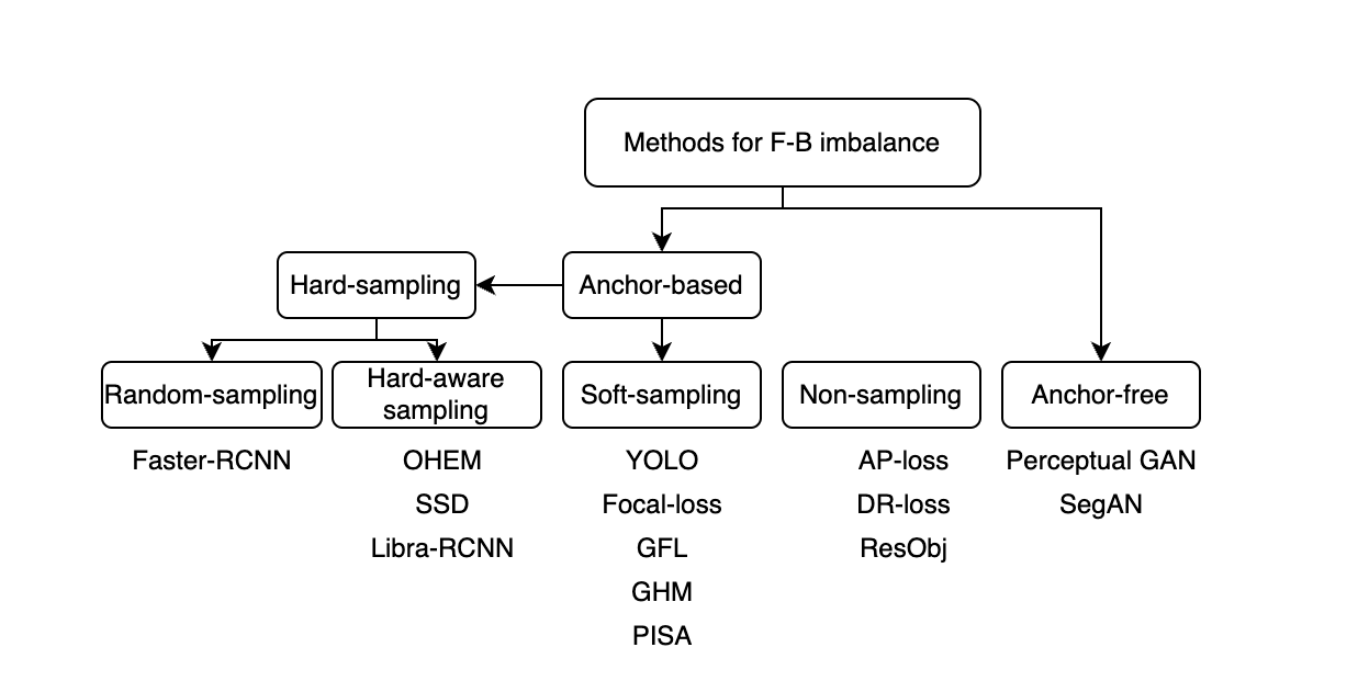}
    \caption{A taxonomy of the methods for handling F-B imbalance.}
    \label{fig:tax}
\end{figure*}

Traditional object detection techniques have two main phases: (1) a region proposal (RP) generator and (2) a feature extraction and classification module. Selective search \cite{uijlings2013selective} or region proposal networks (RPNs) \cite{ren2015faster} are usually used as the first step, and they generate a series of candidate regions, referred to as anchors, in a given image. Some recently proposed single-stage methods, such as SSD \cite{liu2016ssd} and YOLO \cite{jiang2022review,redmon2018yolov3}, sample boxes uniformly from the whole image and eliminate the RP step, and the sampled boxes are also seen as anchors. In addition to these anchor-based detection methods, many anchor-free methods have also been proposed and achieved decent performance in recent years, such as GAN-based \cite{rabbi2020small,han2019synthesizing,li2017perceptual} and anomaly detection-based methods \cite{basharat2008learning,ganokratanaa2020unsupervised,konz2023unsupervised}. 

In the following section, we categorize and discuss strategies for dealing with F-B imbalance in two major scenarios, anchor-based and anchor-free, with a detailed taxonomy shown in Figure \ref{fig:tax}.

\subsection{Anchor-based methods}
As defined in \cite{ren2015faster}, an anchor is labeled as positive (foreground or FG) if the intersection over union (IoU) between it and any ground truth bounding box is above a given threshold, i.e., 0.5. Otherwise, it becomes a negative (background or BG) anchor. Then, the foreground discrimination step can be perceived as a classification task on positive and negative anchors, and the objective of this step can be represented as
\begin{equation}
    L_c = \Sigma w_i C(p_i),
\end{equation}
where $C(p_i)$ is the cross-entropy loss for the proposed anchor $p_i$, and $w_i$ is the contribution of that selected $p_i$.

Due to the F-B imbalance in the original image (as we defined before), these anchor-based methods result in a very small proportion of positive examples on the RP step and a dramatic positive and negative class imbalance. Due to the existence of this unavoidable class imbalance brought by anchor-generating steps, there are several approaches to address the F-B class imbalance, which can be grouped into 3 main categories: (1) hard sampling methods, (2) soft sampling methods, and (3) other methods. We will detail some representative methods in each category below and select some of these methods for a systematic experimental evaluation in Section 4.

\subsubsection{Hard-sampling methods}
Over-sampling and down-sampling \cite{hardle2009variable} are widely applied in traditional classification problems to deal with class imbalance. Due to the massive number of BG samples in object detection, down-sampling is a common strategy for addressing the F-B imbalance problem, also called hard sampling. In hard sampling methods, the weights $w_i$ in $L_c$ are restricted as binary, i.e., 0 or 1, which means only a subset of positive and negative anchors will be sampled towards the classification objective. For example, if the weight for half of the negative samples is set to 0 and all positive ones to 1, the imbalance ratio ($IR_c$) is roughly doubled. 

There are several hard-sampling methods, and the most straightforward one is the \textbf{random-sampling} method, which is commonly used to select examples to meet a certain ratio randomly. The well-known structure incorporating this tactic is the R-CNN family \cite{girshick2014rich,he2015spatial,girshick2015fast}, where a subset of negative samples is selected randomly. To be more specific, the region proposal network (RPN) \cite{ren2015faster} first samples a mini-batch of R anchors (also called the region of interest in the original paper) from N images, and R and N are set as 128 and 2 in \cite{ren2015faster}. Among these anchors, positive samples with a higher IoU ($IoU \geq 0.5$) and negative samples with a lower IoU ($IoU<0.5$) are then randomly selected to meet a final ratio $r$ of 1:3 during the training stage of the detection network, thus ensuring that 25\% of this mini-batch is FG ROIs. Among the R-CNN family, \textit{Faster R-CNN} \cite{ren2015faster} is a relatively representative and widely used method.

Inspired by the success of adaptive boosting \cite{margineantu1997pruning}, where paying more attention to misclassified examples in the next optimization step can improve accuracy, \cite{abney2002bootstrapping} stated that when training the network, hard samples contribute more to updating the model weights (providing a larger loss). Also, due to a large number of negative samples and the intuitive representation difference in foreground and background, most of the BG samples are easily discriminated from the FG samples and classified correctly by the network. Random sampling scheme is easily dominated by simple samples and therefore fails to achieve optimal performance. 
Therefore, there are several hard-aware sampling strategies that are alternatively characterized as tentatively sampling a greater percentage of hard examples. \textit{Online Hard Example Mining (OHEM)} \cite{pang2019libra} is a proposed method that only considers the top B cases in this batch with the largest RoI losses for updating the network during backpropagation. OHEM removes heuristic hyperparameter settings such as the final negative-positive ratio ($r$) and reduces the computation cost by selecting a small number of ROIs for the backward pass.

\textit{Single-Shot Detector (SSD)} \cite{liu2016ssd}, as a single-stage detector, also retrieves a drastic imbalance of positive and negative samples after generating uniformly gridded boxes and performing a box-matching procedure. Similar to OHEM, it ranks the grids based on the confidence loss (negative confidence score) and picks the top grids such that the ratio between positives and negatives is not less than a predetermined ratio, i.e., 1:3.

It was recognized that training OHEM-based algorithms can be unstable when confronted with noisy labels; thus, \textit{Libra-RCNN} \cite{pang2019libra} proposes an IoU-balanced sampling strategy.
It establishes a relationship between the sample's difficulty and the IoU between samples and their associated ground truth, where 60\% of hard negatives overlap larger than 0.05, but only 30\% are acquired using random sampling, as seen in Figure \ref{fig:illustrate} (bottom right).
Thus, Libra-RCNN divides the IoU sampling interval into K bins and uniformly samples negative anchors from each bin, resulting in a larger number of instances with higher IoUs than random sampling (this method can also be seen as hard negatives).
The selected probability for each sample under IoU-balanced sampling is:
\begin{equation}
    p_k = \frac{N}{K}\frac{1}{M_k}, k \in [0,K),
\end{equation}
Where $M_k$ is the number of sampling candidates in the corresponding interval denoted by k, and K is set as 3 in \cite{pang2019libra}.

\subsubsection{Soft-sampling methods}
The distinction between soft and hard sampling methods is that in soft sampling, the weight ($w_i$) is a scalar that presents the relative significance of its corresponding sample, whereas it becomes a binary value in hard sampling. In this manner, each sample contributes towards the final objective in various scales. 

Utilizing different constant coefficients for foreground and background classes is common. As a relatively representative one-stage method, \textit{YOLO} \cite{redmon2018yolov3} uniformly halves each background sample's loss contribution value ($w_i=0.5$).

\textit{Focal-loss}, which was first proposed in one-stage detector Retinanet \cite{lin2017focal}, dynamically adjusts the weights based on the difficulty of the instances by the formulation:
\begin{equation}
    w_i = \alpha (1- p_i)^\lambda,
\end{equation}
where $p_i$ is the confidence score and is negatively related to the weights. $\alpha$ and $\gamma$ are two hyper-parameters. To be noticed, if $\gamma$ is set to 1, Focal-loss then deteriorates to the normal CE-loss, seen as Figure \ref{fig:illustrate} (Focal-loss).

Also, the original Focal-loss algorithm only supports discrete $\{0, 1\}$ category labels for background and foreground samples, which precludes potential application with continuous IoU labels $[0, 1]$ as supervisions and rises instability during training. 
By specifically introducing a Qualify Focal-loss (QFL) for dealing with difficult samples by producing their continuous $[0, 1]$ quality estimations and a Distribution Focal-loss (DFL) for the dense location of bounding boxes, Generalized Focal loss (GFL) \cite{li2020generalized}, as an improvement and replacement version of Focal-loss, could target to any desired continuous value and also to dense objects.

Similar to Focal-loss, the \textit{Gradient Harmonizing Mechanism (GHM)} \cite{li2018gradient} observes that there is an excessive number of samples with a small gradient norm, a restricted number of examples with a medium gradient norm, and a disproportionately high number of samples with a large gradient norm. Due to this disharmony of gradients, the authors believe that a large portion of instances contributes little to the model optimization because of their small gradient norm. These instances are known as relatively easy instances. Thus, GHM-C (the classification part of GHM) penalized the contribution of samples within a similar range of gradient norms by:
\begin{equation}
    w_i = \frac{N}{GD(g_i)},
\end{equation}
where $GD(g)=\frac{1}{l_\epsilon(g)} \sum_{k=1}^N \delta_\epsilon\left(g_k, g\right)$ and $g_i=|p-p^*|$, is the gradient density which means the number of samples in this batch which have a similar gradient norm as $i$, seen as Figure \ref{fig:illustrate} (GHM-C). 

\textit{Prime Sample Attention (PISA)} \cite{cao2019prime} proposes an importance-based sample re-weighting (ISR) that assigns weights to positive and negative examples based on their contribution to the detection performance (i.e., mAP) where examples with a higher impact on mAP metric are favored. More specifically, PISA first builds an IoU-Hierarchical local rank (HLR) $(r1,r2,…,rj)$ where $0 \leq r_i \leq n_{j-1}$ and $n_j$ is the total number of samples for class $j$, to rank the examples for each class (N foreground classes and 1 background class) based on their desired property (i.e., IoU for positives and classification score for negatives) and sample $i$ is weighted as follows:
\begin{equation}
\begin{aligned}
    u_i &= \frac{n_{r_i}}{n_{max}} \\
    w_i &= ((1-\beta)u_i + \beta)^\lambda,
\end{aligned}
\end{equation}
where $n_{max}$ is the max value of $n_j$, which ensures the samples at the same rank of different classes will be assigned the same value $u$. $\gamma$ is the degree factor similar to that in focal loss, and $\beta$ is a bias that decides the minimum sample weight. Note that these parameters are calculated on positive and negative samples  separately, and $w_i$ increases when samples have a higher desired property. Also, besides re-weighting the classification loss by ISR, PISA proposes a Classification-Aware Regression Loss (CARL). By jointly optimizing the box regression and classification sub-branches, CARL can further boost the scores of prime samples, seen as Figure \ref{fig:illustrate} (Prime samples).

\subsubsection{Other methods}
In general, the methods discussed in the previous section are modified with respect to the classification loss $L_c$. Alternatives to hand-crafted sampling heuristics have evolved to reduce the number of hyper-parameters during training and to preserve the sample distribution throughout training and interference. Some of these techniques opted to replace the previous classification loss $L_c$. \textit{Average-precision loss (AP loss)} \cite{Chen_2021} is proposed for a ranking problem to optimize average precision as a substitute for the classification task in one-stage detectors. The authors claimed that AP could better eliminate the impact caused by true negatives introduced by background. Similarly, \textit{Distributional Ranking loss (DR loss)} \cite{qian2020dr} chooses to rank the expectations of the derived distributions of foreground and background and to emphasize a large margin at the decision boundary by forcing the estimated probability of positive examples to be higher than that of  negatives by a margin:
\begin{equation}
\min _\theta \ell\left(\max _{j_{-}}\left\{p_{j_{-}}\right\}-\min _{j_{+}}\left\{p_{j_{+}}\right\}+\gamma\right),
\end{equation}
where $p$ denotes the probability and $j+$ and $j-$ denote (positive) foreground and (negative) background examples, and $\gamma$ is a non-negative constant.

On the other hand, some methods have been proposed to substitute the sampling heuristic. \cite{chen2019residual} proposed a \textit{Residual Objectness (ResObj)} mechanism as a new branch to deal with F-B imbalance by an end-to-end optimization. Instead of using hand-crafted sampling or reweighting, they reformulate this new branch to predict the residual objectness scores under a cascaded refinement procedure. This new branch can be easily incorporated into any one-stage/two-stage detectors, such as YOLO \cite{redmon2018yolov3} and Faster-RCNN \cite{ren2015faster}, and the original classification branch could follow the vanilla classification loss $L_c$. During inference, class-specific scores are obtained by multiplying classification and objectness branch outputs. 

\subsection{Anchor-free methods}
Recently, several generative methods have emerged to handle the F-B imbalance problem. \textit{Perceptual GAN} \cite{li2017perceptual}, which intentively lifts representations of small objects to ``super-resolved'' ones, alleviates the F-B imbalance since the small objects archive similar characteristics as large objects. \textit{SegAN} \cite{zhang2018seggan} combines the generator with a segmentation network as a detector along with a multi-scale L1 loss and gets around the F-B imbalance by patch-level training. Similarly, \cite{konz2023unsupervised} used patch-level image completion to detect breast tumors in digital breast tomosynthesis much smaller than the full scan image size. Although some promising trials have been conducted, these GAN-based methods are more complex and have not yet established themselves as a more common method in the object detection field.

On the other hand, some strategies can help with F-B imbalance-related problems. Such as, for an extreme F-B imbalanced dataset, the object sizes vary greatly. Thus, the predefined Feature-extractor might not be capable of grasping extremely small objects. \textit{Feature Primarid network (FPN)} can be an effective way to address this concern. This technique has been employed before it was formally introduced. RetinaNet \cite{lin2017focal}, which was introduced in Focal-loss, can be viewed as Resnet plus the FPN. Since this module is not specifically designed for the F-B imbalance and can be applied to the majority of backbone architectures, it is excluded as one individual strategy from the discussion.

\subsection{Remaining issues for these methods}
All of the methods previously mentioned suggest various ways to reduce F-B imbalances, and they have demonstrated improved performance on a few particular public datasets. However, these methods are conducted on these complicated benchmark datasets in which the size of targets varies, F-B imbalances come from various sources, and F-B imbalances are under various intensities. For example, the average size of targets in COCO ranges from 0 to 1000 pixels, seen in Figure \ref{fig:coco-fig}. Despite the claims that they have improved performance, it is obscure how each method has improved for the F-B imbalance alone. As a result, when working with a custom dataset, we might get less assistance and references and feel it challenging to choose which approach to employ. Therefore, we conduct a series of experiments to show different perspectives of F-B imbalance and how the above-mentioned methods address each.   

\section{Selected methods and data}
\subsection{Methods selected for this study and training details}
\subsubsection{Selection of methods and implementation details}
For the experiments, we chose 9 representative object detection methods from those described above: Yolo-v3, Faster-RCNN, GHM, PISA, SSD, OHEM, Libra RCNN, Focal loss, and Generalized Focal Loss (GFL), as seen in Table \ref{tab: methods}. There were two criteria for the selection: (1) having strategies for addressing F-B imbalance and publicly available official implementation codes, or (2) being frequently used for tasks involving object detection. In our selection, Faster-RCNN is the widely implemented two-stage backbone for object detection, while OHEM and PISA are two popular components added to the normal baseline. SSD and YOLO-v3 are two representative 1-stage detection algorithms. Finally, Focal-loss, GHM, and GFL are within the modified loss function family. The details of implementing these methods are shown in Table \ref{tab: methods}. 

\subsubsection{Model training details}
Each method is trained for 36 epochs and evaluated at the last checkpoint. Learning rates are set to 0.0025, except SSD, which is set to $10^{-3}$ as in their original setting because of its convergence speed. All models are trained on GPU GTX3090. Each experiment is repeated 10 times with different random seeds ranging from 0 to 9 to obtain trustworthy and reproducible results.

\begin{table*}[]
\caption{The detailed descriptions of the implemented methods for F-B imbalance in this work.}
\label{tab: methods}
\scriptsize
\begin{tabularx}{\textwidth}{llXXXl}
\hline
Method     & Stage Num              & Base network backbone we employed      & Strategies for F-B imbalance                                                                      & Sampling category                                      & Year              \\
\hline 
Faster-RCNN \cite{ren2015faster}                              & Two-stage                  & Resnet 50                  & Randomly down-sample  background samples                                                                                               & Random sampling &    2015               \\
SSD-300   \cite{liu2016ssd}            & One-stage & Vgg-16    & Only choose  background samples with max loss contribution   & Hard sampling   & 2016 \\
OHEM  \cite{shrivastava2016training} & Two-stage                  & Resnet 50                  & Only choose pos\&neg   with max-loss contribution                                          & Hard sampling                                         &    2016               \\
Libra-RCNN   \cite{pang2019libra}                                     & Two-stage                  & Resnet 50                  & IOU balanced sampling                                                                             & Hard sampling                                         &    2019               \\
Focal-loss \cite{lin2017focal}                                     & One-stage                  & RetinaNet                  & Promote examples with larger loss                                                                 & Soft sampling                                         &    2017               \\
GHM \cite{li2018gradient}     & One-stage                  & RetinaNet                  & Promote examples with larger loss   by suppressing effects of outliers                            & Soft sampling                                         &        2018           \\
PISA  \cite{cao2020prime}                                           & Two-stage                  & Resnet 50                  & Ranks the examples for each class   based on their desired property & Soft sampling                                         &   2020                \\
YOLO   (v3) \cite{redmon2018yolov3}                                      & One-stage                  & DarkNet 53                 & Constant $w_i$ =0.5 for background   examples                                                      & Soft sampling                                         &      2018            \\
GFL  \cite{li2020generalized}        & One-stage                  & RetinaNet & Combine Quality Focal Loss (QFL)   and Distribution Focal Loss (DFL)                              & Soft sampling                                         &  2020 \\
\hline
\end{tabularx}
\end{table*}

\subsection{Datasets}
We started our experiments with a balloon dataset which we synthesized in order to control object size and density freely, and three real datasets from different domains: the most widely used benchmark COCO dataset for object detection tasks, an air-drone dataset with small object size but relatively dense distribution, and lastly a medical breast dataset with not only small object size but also very sparse object distribution. Below we describe these datasets in detail.

\subsubsection{Balloon dataset}
As we discussed before, the F-B imbalance ratio (IR) is mainly influenced by the object size and the number of objects per image. To control the parameters of our experiments and manage the IR quantitatively in our analysis, we generated several synthetic balloon datasets as the first stage of our experiments. We collected a background subset with 100 natural images (w × h) and an object subset with 50 different balloons online. All the background images are resized to $800\times1000$ for $w \leq h$ or $1000\times800$ for $w > h$, respectively, with their original width and height. By randomly resizing, rotating, injecting, and positioning balloons into these backgrounds, we can generate several datasets based on various image properties. Specifically, denote $N$ as the number of images, M as the number of objects (balloon) per image, and $S$ as the range of the object size ($S_{obj}$) for this dataset ($ \forall S_{obj} \in [S, 2S)$); we can present a subset as (Img-N-balloon-$M_c$-size-$S_c$). Also, when the density of objects on each image increases, occlusions can occur, but we ensured that the occlusion between objects does not exceed 90\% area of the object.

\subsubsection{COCO dataset}
The COCO dataset \cite{lin2014microsoft} is the most widely used object detection dataset. It includes 82k training images and 40k validation images for a total of 80 different objects, making it a very large-scale dataset. Because it primarily consists of scenes from everyday life and contains a lot of objects with very small sizes, the coco dataset also has a significant F-B imbalance for some categories. The imbalance ratio (IR) is shown in Figure \ref{fig:IR}, and we can see that for more than half of the object categories, the IR is less than $10^{-2.5}$.

\subsubsection{VisDrone	dataset	– dense	object	dataset}
VisDrone2019-Detection \cite{9573394} consists of images captured by Unmanned Aerial Vehicles (UAV) platforms at various locations and altitudes. The collection includes predominantly small, densely distributed, and partially obscured items. More than 540,000 targets’ bounding boxes are marked with 11 predefined categories, including walkers, persons, bicycles, cars, vans, trucks, tricycles, awning vehicles, buses, and motors. The total dataset contains 6471 training data and 548 validation data.

\subsubsection{Duke BCS-DBT dataset – sparse object dataset}
In real applications, the F-B imbalance is also common in medical images because the RoIs, i.e., lesions, are often very small. Moreover, images containing RoIs only comprise a small proportion of the overall dataset due to the natural rarity of diseases. The Duke BCS-DBT dataset is a challenging digital breast tomosynthesis dataset \cite{konz2023competition} for breast tumor detection. This dataset is severely affected by the F-B imbalance because the size of the objects (tumors) is extremely small compared to the large high-resolution background area, and the number of tumors is also very small for the entire dataset, where the dataset contains 5610 images. In contrast, only 274 of them contain tumors.

\subsection{Evaluation Metrics}
\textbf{mAP}. We selected the commonly used evaluation metric of average precision (mAP) to present our results. Given an image with ground truths bounding boxes (bboxs) and class labels, each proposed prediction is considered a true positive (TP) when (1) the IoU between it and its nearest ground truth bbox is greater than some threshold $\theta$ and (2) the predicted label matches the ground-truth class label. Predictions that do not satisfy the above conditions are false positives (FP), and the ground truths without any corresponding predictions are false negatives (FN). Thus, we can obtain a precision-recall curve by stepping the IoU threshold $\theta$ from a certain range and calculating the area under this curve as average precision (AP). The mAP is the class-wise average of AP across all the object classes:
\begin{equation}
    mAP = \frac{1}{N}\Sigma_{i=1}^N AP_i,
\end{equation}
where N is the number of classes. The mAP takes into account the trade-off between precision and recall and is a widely used metric for evaluating detection performance. Throughout the experiments, we report mAP50 where $\theta=0.5$ and mAP(50:95) where $\theta$ ranges from 0.5 to 0.95 with step size 0.05, and we simplify it as mAP later.

\textbf{Model Size}. We measured the size of these models by counting the parameters inside the model that requires gradients.

\textbf{Training and inference time}. The training and inference times are calculated as follows: (1) the total time to complete training epochs; and (2) the time to evaluate a single image. The training time is measured on a single GTX 3090, and the inference is conducted under the CPU setting.

\section{Experiments}

We designed various experiments to evaluate the impact of F-B imbalance from various perspectives, such as different object sizes, different data sets, different sources of F-B imbalance, and the performance of various methods, to achieve a multi-faceted understanding of the F-B imbalance problem.
F-B imbalance ratio for object type c, $IR_c \sim f(M_c, S_c)$ is mainly controlled by both the object size and the average object number per image. We focused on these specific questions:
\begin{enumerate}
    \item How does the object size as the main source of F-B imbalance influence the object detection performance?
    \item How does the number of objects as the main source of F-B imbalance influences the object detection performance?
    \item How does the imbalance ratio $IR_c \sim f(M_c, S_c)$ influence the object detection performance?
    \item Which methods are the best at addressing the F-B imbalance problem?
\end{enumerate}

\subsection{Impact of object size on model performance}
\subsubsection{Experimental design} 
\textbf{On balloon dataset}. We selected training data of varying dataset sizes, including 50, 100, 200, 500, etc. For each given size of the training dataset, we selected balloon objects with varying average sizes: 16-32 (S=16), 32-64 (S=32), ..., 128-256 (S=128) on average, with each image containing one object inserted (Img-N-balloon-1-size-S). We utilized the selected 9 methods to address the F-B imbalance on these various training datasets. For each training set, we also prepared 200 unused test images that were of the same size as the training objects for evaluation. For instance, we trained one model on a set of 100 images of balloons with an object size of 16 (Img-100-balloon-1-size-16) and then evaluated it on a test set of 200 images of size 16, each image containing a single item (Img-200-balloon-1-size-16).

In this particular scenario, it is evident that the imbalance ratio $IR=f(S_c\mid  M_C=1)$ for each balloon training dataset is mainly determined by the object size parameter $S_c$, while maintaining the number of the presence of objects per image ($M_C=1$) constant. Moreover, it can be inferred that $k^2IR_c\sim f(kS_c| M_C=1)$, where $k$ denotes the scaling factor applied to the object size. For instance, if all target objects in the dataset are doubled in size, the resulting imbalance ratio of this dataset is multiplied by a factor of four.

\begin{figure*}
    \centering
    \includegraphics[width=0.8\columnwidth]{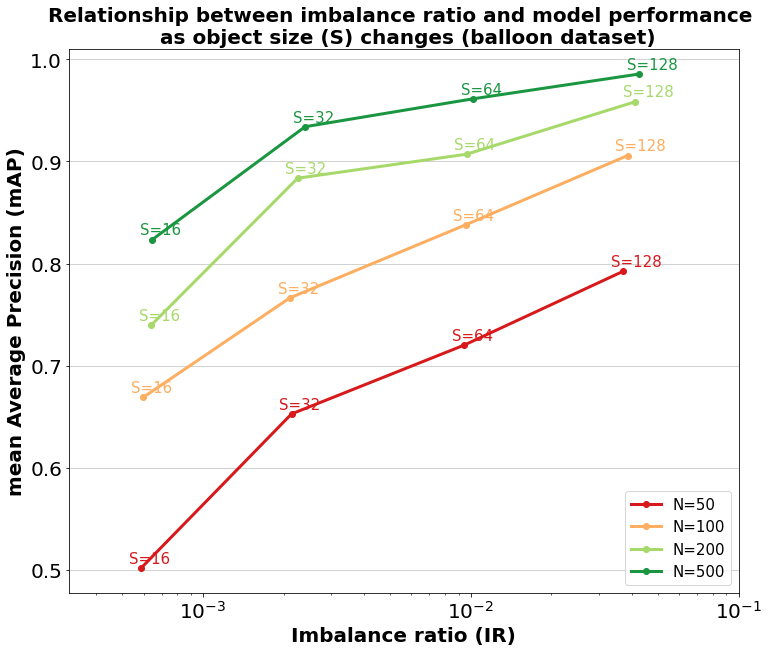}
    \caption{The relationship between foreground-background imbalance ratio (IR) of the training set and the mean average precision (mAP) on the evaluation set, which is reported by the maximum mAP among all the selected methods, when the object size is the only influencing factor of the imbalance ratio (IR) on balloon dataset. N-50, 100, 200, and 500 denote the different sizes of the training dataset. S-16, 32, 64, and 128 denote the average object size in the dataset.}
    \label{fig:size-results}
\end{figure*}

In an effort to assess the influence of object size on detection performance, independent of the choice of methodology, we conducted a calculation of the maximum mAP (mean average precision), herein referred to as mAP+. This value represents the observed upper limit of detection performance for the specific dataset under consideration. The results of this experimental endeavor are illustrated in Figure \ref{fig:size-results}.

\begin{figure*}
    \centering
    \includegraphics[width=0.8\columnwidth]{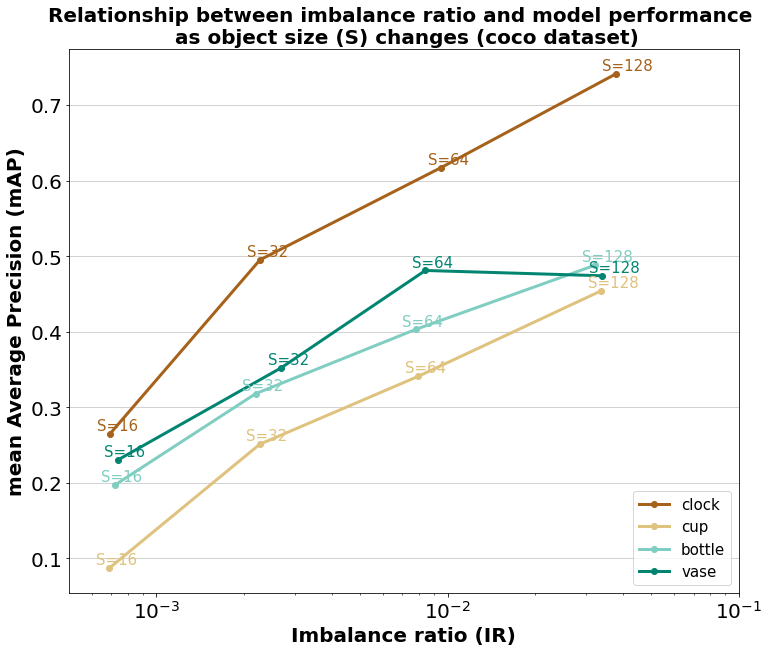}
    \caption{The relationship between foreground-background imbalance ratio (IR) and upper-bound of detection performance when the object size is the only influencing factor for the coco dataset. Four different object categories are selected for experiments, and each training set size is 200.}
    \label{fig:coco-size-results}
\end{figure*}

\textbf{On coco dataset}.
To study the relationship between object size and detection performance on COCO, we performed a preprocessing step to generate subsets of coco, which involved selecting our target object and creating a subset by narrowing a particular object size range. First, we resized the images into $800 \times 1000 (w \leq h)$ pixels or $1000 \times 800 (w > h)$ pixels, respectively. Then we picked four relatively tiny object categories from the overall training data set: clock, cup, bottle, and vase. The selected categories are further partitioned into smaller training data sets of S=16, S=32, S=64, and S=128 according to their size ranges. We also excluded images with more than one target object, and thus the average object number (M) is controlled as 1 for these datasets. Using a similar notation, Img-200-clock-1-size-32 means we select 200 images with a single clock on each image with sizes ranging from 32 to 64. 
Under each coco subset, the imbalance ratio (IR) is entirely controlled by the item size (S) due to the restriction of having only one object in each image. Thus, the previously described issue with the complexity of the COCO dataset has been overcome, and the detection performance can be easily correlated to the F-B imbalance $IR \sim f(S_c|M_c=1)$. 
The experimental results for each subset with respect to the upper bound of detection performance (mAP+) are shown in Figure \ref{fig:coco-size-results}. 

\subsubsection{Experimental results} Figure \ref{fig:size-results} presents an overview of the results obtained by varying the object size on the balloon dataset. Notably, it can be observed that the observed upper bound of mean average precision (mAP+) increases in proportion to the increase in the imbalance ratio (IR) and the average object size across all dataset sizes. This confirms our initial hypothesis that object size positively correlates with detection performance. Additionally, we found that the increase in performance due to the change in object size is more prominent between objects S=16 and S=32, as compared to S=64 and S=128. Specifically, mAP+ improves from S=16 to S=32 by 0.151, 0.097, 0.144, and 0.110 for the N-50, N-100, N-200, and N-500 datasets, respectively, and from S=64 to S=128 by 0.07, 0.06, 0.05, and 0.02, respectively. This suggests that modifying the object size has a more substantial effect on detection performance for smaller objects than larger ones. Furthermore, upon analyzing the growth of each curve from N=50 to N=500, we observed that the mAP growth from the smallest object size to the largest is 0.291, 0.237, 0.219, and 0.162, respectively. This indicates that networks benefit more from increasing object size when the number of training images is limited, and IR has a greater influence on smaller training sets.

Figure \ref{fig:coco-size-results} presents the performance of aggregated from various methods on the COCO subsets. We observed an overall improvement in detection performance as the object size and IR increased. However, we noted that these methods performed worse in the category ``vase'' when the object size increased from S=64 to S=128. This can be attributed to the limited availability of validation data, as we only collected 18 images for the S=128 validation set, which may not be representative enough. Furthermore, we found that the detection performance varied considerably for different object categories with similar sizes of objects. For instance, when the object size increased from S=32 to S=128, the mAP+ improved from 0.251 to 0.454 for the category ``cup'' and from 0.494 to 0.741 for the category ``clock''. This highlights that the category and inherent representation properties of the object also play a crucial role in determining detection performance, in addition to the IR.

\subsection{Impact of number of objects on model performance}
\subsubsection{Experimental design} 
\textbf{On the synthesis balloon dataset.} Considering other real-world datasets, it is hard to adjust the average number of objects, as noted as object density, with the fixed dataset size; we conducted our experiments on the number of objects on the synthetic balloon dataset. In the balloon dataset, we selected training data sizes 50, 100, 200, and 500 to evaluate the impact of the number of objects on detection performance for various dataset sizes. We set the object size to S as 32 and adjusted the number of objects inserted in each image. For any dataset size N, the average number of objects per image (M) increases from 1, 2, 4, and 8 to 32. Under this setting, for each balloon training dataset, $IR=f(M_c\mid S_c)$ is only controlled by the average number of objects $M_c$. Also, it could be inferred that $kIR~f(kM_c|S_c)$, where $k$ is the scaling factor of the number of objects. On these datasets, we implemented experiments on our selected 9 methods. For testing, we used the non-overlapping 200 images, each with one balloon of size $S_c=32$ inserted.

\begin{figure*}
    \centering
    \includegraphics[width=0.8\columnwidth]{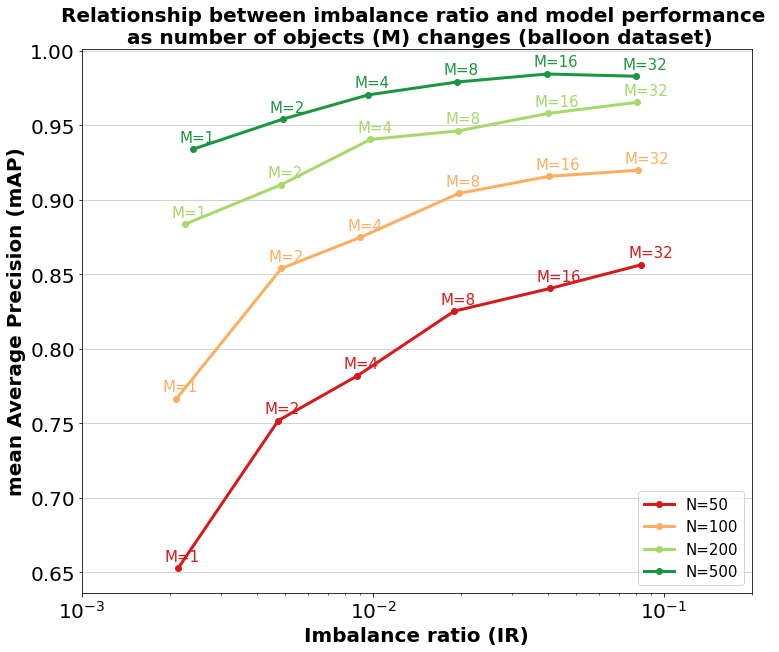}
    \caption{The relationship between foreground-background imbalance ratio (IR) and upper-bound of detection performance when the object size is the only influencing factor for the coco dataset. Four different object categories are selected for experiments, and each training set size is 200.}
    \label{fig:number-results}
\end{figure*}

\subsubsection{Experimental results}
Figure \ref{fig:number-results} shows the detection performance with respect to the number of object changes. We observed that these curves generally increase when object density M (average number of objects per image) grows. And we noted that the trend of the increasing detection performance with IR’s growth was more significant when the dataset size was small. When N is 50, as the density increases from M=1 to M=32, the mAP+ across all methods changes from 0.6528, 0.7519, 0.7819, 0.8253, 0.8406, and 0.8565. In comparison, the mAP+ with respect to the density growing is 0.934, 0.9542, 0.9705, 0.979, 0.9845, and 0.983 when N is equal to 500. For a dataset of 50 balloon images, the increase was 31.2\%, but for a dataset of 500 balloon images, it was only 5.2\%.
We also observed the mAP+ increase plateaued at the larger side of IR and displayed some subtle drop from S=16 to S=32. Specifically, mAP+ drops 0.002 for the N=500 balloon dataset. This can be caused by the fact that a higher object density would increase the likelihood of overlap and occlusion of target objects, making training more difficult.

\subsection{Joint impact of object size and number of objects on model performance}
As we discussed in the previous two sections, object size and the number of objects per image both play a significant role in influencing detection performance. In this section, we analyzed the joint impact of the number of objects and object size on model performance.

\subsubsection{Experimental design}
In this section, we evaluated a more complex scenario wherein both $S_c$ and $M_c$ are altered, leading to a variation in the size of the objects from S=16 to S=128 and in the number of objects ranging from M=1 to M=64. Within this parameter space, the training datasets demonstrated a spectrum of foreground-background imbalances. At one extreme, each image had a pronounced imbalance with a solitary diminutive object (S=16, M=1). Conversely, at the other extreme, the dataset predominantly consisted of images with a considerably lesser concern regarding the foreground-background imbalance, as they contained 64 substantial objects (S=128, M=64) each.

We tried simultaneously increasing object size while reducing object density to keep a relatively stable IR. We discovered empirically that doubling the object size and quadrupling the object number similarly impact $IR$. Specifically,
\begin{equation}
    IR \sim f(S_c,M_c) \simeq f(2S_c,\frac{1}{4}M_c).
\end{equation}
As a result, we could multiply the imbalance ratio by a factor of 4 in two ways: one is to double the object sizes, and another is to quadruple the number of objects (seen in Figure \ref{fig:ir-results}, right of the arrow). The synthetic balloon dataset used in these experiments consists of 200 images as the training set (Img-200-balloon-$M_c$-size-$S_c$), and the 200 non-overlapping images with only one object injected were used for testing (Img-200-balloon-1-size-$S_c$).

\subsubsection{Experimental results}
\begin{figure*}
    \centering
    \includegraphics[width=\columnwidth]{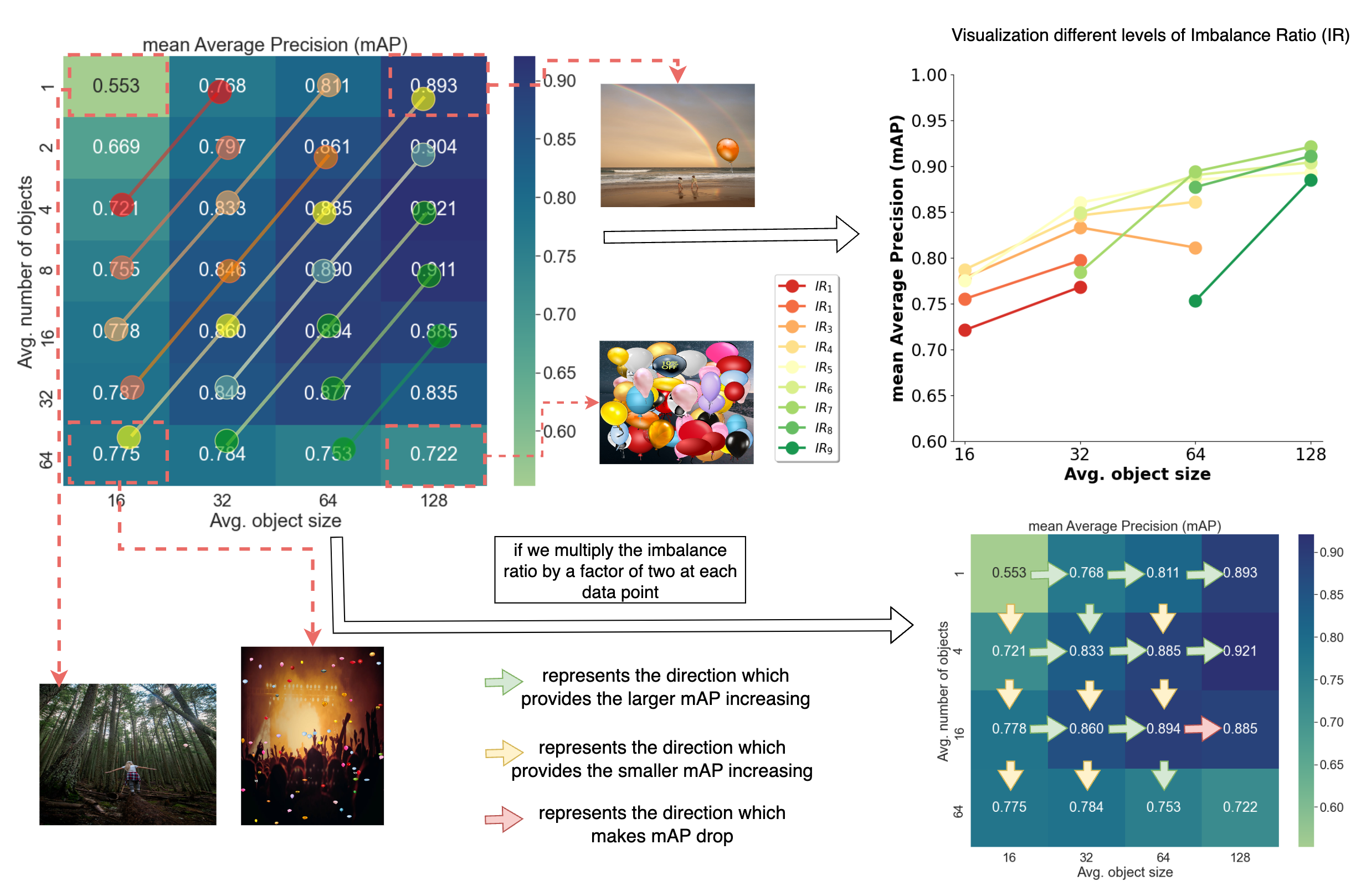}
     \caption{The correlation between the foreground-background imbalance ratio (IR) and the average detection performance of various methodologies as the sizes and the number of objects vary. Each curve depicts a distinct intensity of foreground-background imbalance, with the red curve ($IR_1$) signifying the most extreme imbalance, and the green curve ($IR_9$) symbolizing the most minimal imbalance.}
    \label{fig:ir-results}
\end{figure*}

Figure \ref{fig:ir-results} (left) illustrates the results for several combinations of object sizes and the number of objects, ranging from extremely small and sparse objects in the table's upper-left corner to large and dense objects in the right-bottom corner. We found that, when we increase IR by a determined ratio (saying 4), 8/9 of the chances, the performance will increase, until there is extreme overlapping of objects. Also, 7/8 of the chance, the higher increase is achieved by an increase in the object size, which reflects that increasing  object sizes are the preferable solution for increasing detection performance. These experiments also revealed that if there are numerous large objects, like in dataset Img-200-balloon-64-S-128, the average detection performance (mAP) is respectable at 0.722, which is much smaller than datasets with fewer objects, such as Img-200-ballon-32-S-128 at 0.835 even the previous dataset under a smaller F-B imbalance issue.

\subsection{Evaluating the performance of different approaches for addressing foreground-background imbalance}
\subsubsection{Experimental design} Firstly, we operated experiments with respect to each individual method on the different settings of synthesis balloon datasets, these experiments are under the same setting as Section 5.1.1 and Section 5.2.1, but we emphasized each approach’s performance instead of an aggregated performance respect to IR. The results are visualized in Figure \ref{fig:method-results} with mean mAP and standard deviation on each training subset. Similarly, we also visualized the results according to each method on the coco dataset discussed in Section 5.1.1, and the results are shown in Figure \ref{fig:method-coco}. In order to expand the findings for the selection of approaches in the real world, we also provided experiments on two real-world datasets: the \textit{Visdrone dataset} with a dense distribution of small-sized objects and the \textit{Duke BCS-DBT dataset} with a very sparse distribution of small-sized ``tumor'' objects. For the Vis-drone dataset, we calculated the IR based on different categories. The models are trained and evaluated on the original test dataset with publicly available annotations. We also supplied the model size and inference time for each technique in order to provide comprehensive information for method selection in real-world scenarios. 

For the Duke BCS-DBT dataset, we collected the images by slices and did a slice-based 2D detection. Besides training on the original dataset (tumor all) with various tumor sizes, we separated this dataset into two subsets for additional experiments. After resizing the images into 1000 × 800, we split the images into two groups with similar numbers of objects: “tumor small” and ``tumor large''. The “tumor small” subset contains 137 objects with a size smaller than 83 pixels, while the “tumor large” subset contains the remaining 139 objects.

\subsubsection{Experimental results}
\begin{figure*}
    \centering
    \includegraphics[width=\columnwidth]{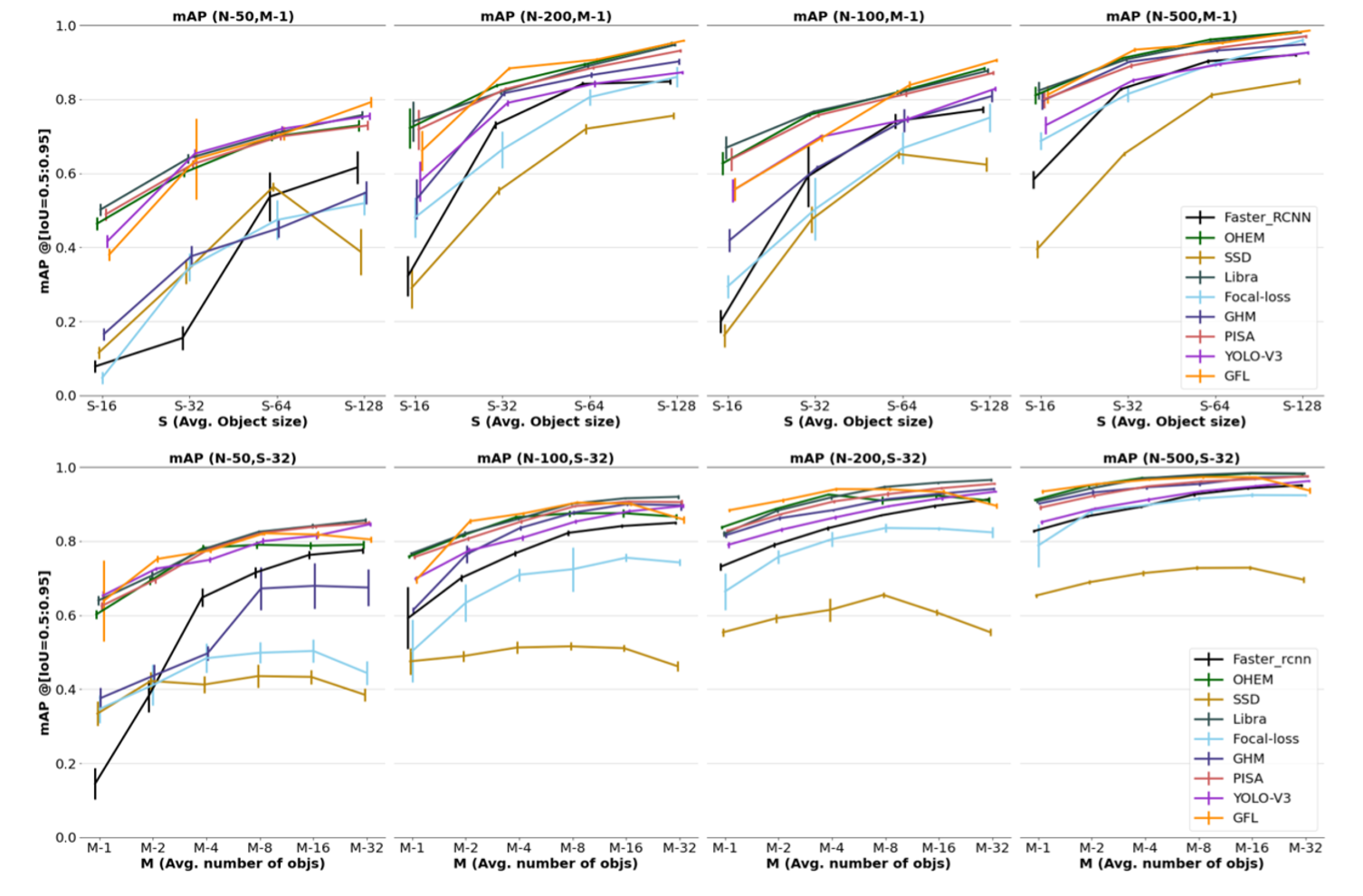}
    \caption{performance of different methods on the synthesis balloon datasets. The first row shows the results when object size changes along the x-axis, and the second row shows the results when the number of objects changes along the x-axis. For each subfigure, N denotes the number of training images, M denotes the average number of objects, and S denotes the average size of objects. Each experiment is repeated 10 times, and the mean mAP and standard deviation of a method are shown as a single data point and the vertical error bar, respectively.}
    \label{fig:method-results}
\end{figure*}

\begin{figure*}
    \centering
    \includegraphics[width=\columnwidth]{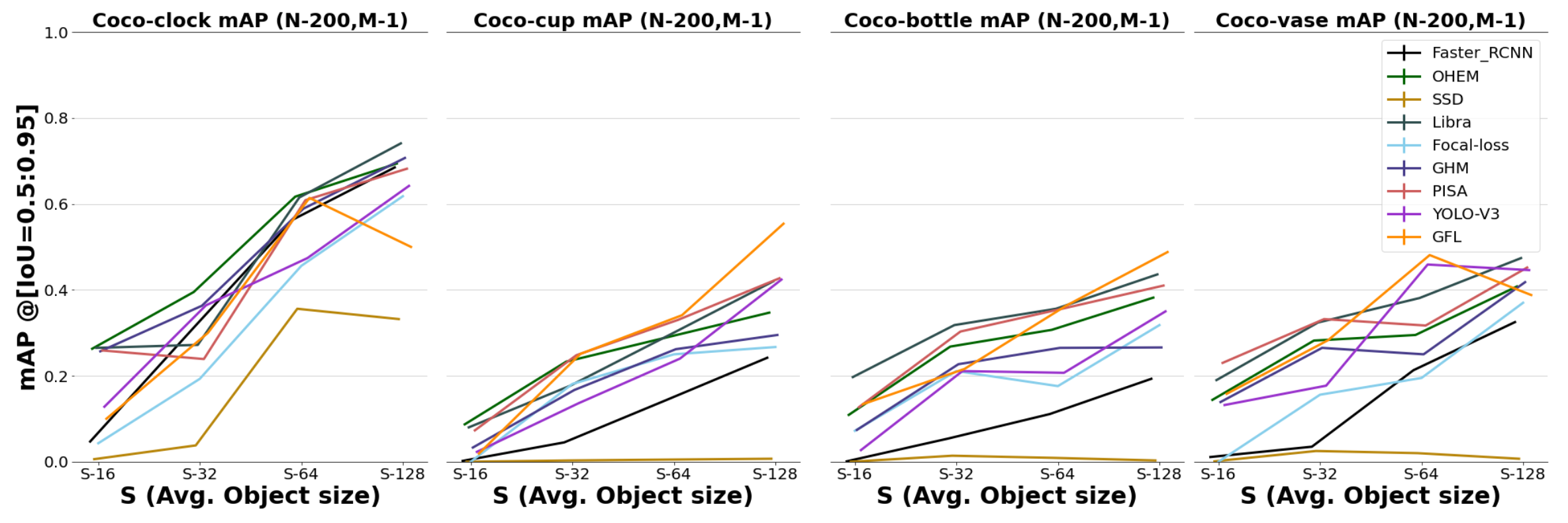}
    \caption{performance of different methods on the coco datasets. It shows the results when object size changes along the x-axis for four different object categories ``clock'', ``cup'' ``bottle'' and ``vase'', respectively. For each subfigure, N denotes the number of training images, M denotes the average number of objects, and S denotes the average size of objects. Each experiment is repeated 3 times, and the average mAP is shown as each single data point.}
    \label{fig:method-coco}
\end{figure*}

For the balloon datasets, shown in Figure \ref{fig:method-results}, we found that the performance of different selections of methods varies a lot when the dataset size is small. For example, the methods for datasets in Img-50-balloon-1-size-S (subfigure row 1, column 1) split into two groups, with a superior group (PISA, YOLO-v3, OHEM, Libra) having an average mAP twice that of an inferior group of methods (Faster-RCNN, GHM, SSD, Focal-loss). It is also similar in datasets in Img-50-balloon-M-size-32 (subfigure row 2, column 1), where the optimal methods GFL, OHEM, PISA, and Libra-RCNN also achieve around 100\% improvement of mAP compared with Focal-loss and SSD.

In row 1, we observe that the standard deviation (vertical error bars in Figure \ref{fig:method-results}) is greater for 1-stage approaches GHM, GFL, and Focal-loss when the training dataset size is small. It achieves a GFL of 0.2  and a Focal-loss of 0.156 on the Img-50-balloon-1-size-32 dataset.

As seen in row 2, the standard deviation for these 1-stage approaches is also greater, with GHM achieving 0.132 and Focal-loss achieving 0.101 for the Img-50-balloon-8-size-32 dataset. As illustrated by the continuously diminishing standard deviation versus dataset size, all algorithms perform more consistently as dataset size. Also seen in row 2, when the average number of objects is extremely large, from M=16 to M=32, although most methods achieve comparable performance, other methods, such as SSD, Focal loss, and GFL, experience a significant decrease in performance, indicating a vulnerability of these methods when occlusions occur.

Regarding the coco dataset, as depicted in Figure \ref{fig:method-coco}, the detection performance is significantly influenced by the selection of the method. For instance, under the setting of extremely small object size of 16, specifically for the category "clock", the detection performance exhibits a substantial variation, ranging from 0.006 for SSD to 0.265 for Libra-RCNN. Similarly, in the "bottle" category, the detection performance spans from 0.001 for Faster-RCNN to 0.197 for Libra-RCNN. Despite this considerable variation across categories and chosen methods, Libra-RCNN, PISA, and OHEM demonstrate commendable performance under conditions of extreme foreground-background imbalance.

\begin{table*}[]
\caption{Results on a real-world \textit{DBT-DCS} datasets, \textit{tumor small}, \textit{tumor large} and \textit{tumor all}. $+-$ represents the standard deviation.}
\label{tab: dbt}
\small
\begin{tabularx}{\textwidth}{X|XX|XX|XX|XXX}
\hline 
Datasets                       & \multicolumn{2}{c|}{Tumor small}   & \multicolumn{2}{c|}{Tumor large}   & \multicolumn{2}{c|}{Tumor all}                     & \multicolumn{3}{c}{Others}            \\
\hline
Metrics                & AP           & AP50         & AP           & AP50         & AP           & AP50         & Train time(s) & params (Num) & Infer time \\
\hline
Faster-RCNN (C4)       & 0.003+-0.003 & 0.038+-0.003 & 0.009+-0.003 & 0.031+-0.005 & 0.037+-0.005 & 0.117+-0.026 & 1122.72      & 32760k               & 0.107          \\
Faster-RCNN (RPN) & 0.006+-0.01  & 0.214+-0.023 & 0.100+-0.01  & 0.274+-0.006      & 0.149+-0.001 & 0.420+-0.007 & 571.75       & 41123k               & 0.117          \\
OHEM                   & 0.051+-0.006 & 0.183+-0.047 & 0.104+-0.002 & 0.291+-0.024 & 0.174+-0.006 & 0.477+-0.021 & 556.22        & 41123k               & 0.107          \\
Libra-RCNN             & 0.058+-0.005 & 0.227+-0.029 & 0.098+-0.013 & 0.250+-0.026 & 0.146+-0.009 & 0.415+-0.027 & 822.2         & 41386k               & 0.058          \\
SSD-300            & 0.027+-0.008 & 0.088+-0.034 & 0.048+-0.019 & 0.143+-0.045 & 0.046+-0.020 & 0.167+-0.033 & 604.79        & 23745k               & 0.015          \\
Focal-loss             & 0.021+-0.010 & 0.122+-0.039 & 0.086+-0.016 & 0.258+-0.026 & 0.086+-0.007 & 0.310+-0.034 & 497.01        & 36104k               & 0.049          \\
GHM                    & 0.055+-0.003 & 0.204+-0.023 & 0.108+-0.010 & 0.293+-0.033 & 0.156+-0.024 & 0.43+-0.032  & 527.33        & 36104k               & 0.064          \\
PISA                   & 0.054+-0.006 & 0.195+-0.019 & 0.108+-0.007 & 0.283+-0.032 & 0.160+-0.008 & 0.417+-0.038 & 622.59        & 41123k               & 0.17           \\
YOLO-v3                & 0.050+-0.006 & 0.145+-0.014 & 0.062+-0.010 & 0.180+-0.016 & 0.108+-0.018 & 0.284+-0.046 & 10331.3       & 61524k               & 0.035          \\
GFL                    & 0.025+-0.019 & 0.103+-0.082 & 0.085+-0.013 & 0.256+-0.021 & 0.097+-0.018 & 0.299+-0.076 & 341.39        & 32033k               & 0.073       \\
\hline 
\end{tabularx}
\end{table*}

The Duke BCS-DBT dataset’s IR of $174.2^{-1}$ indicates a significant F-B imbalance. 
We can see from Table \ref{tab: dbt} that mAP is lower compared with previous experiments and that this dataset is, in fact, much more challenging than the earlier synthetic data and public coco data. 
The algorithms’ performances on the tumor all dataset also vary greatly. The best method OHEM achieves 0.174 on mAP, while the worst Faster-RCNN-C4 only achieves 0.037. The best performance of AP50 is 0.447, while the worst is only 0.117. 
On the tumor all dataset, FPN plays a significant role in improving the detection performance, which increases the Faster-RCNN method from 0.117 of AP50 to 0.420 of AP50. Also, though the Libra-RCNN method didn’t achieve the best performance on the tumor all dataset, it showed the best ability to combat F-B imbalance, which has the best performance on the tumor small subset as 0.058 for mAP and 0.227 for AP50. Additionally, though OHEM has the best performance on the overall dataset, its performance declines for small tumors compared to large ones, showing a relatively lower ability to mitigate the F-B imbalance issue. This also supported our statement in the previous section's assumption that better performance on a complicated real dataset is not enough to demonstrate one strategy’s effectiveness on F-B imbalance.

\begin{figure*}
    \centering
    \includegraphics[width=\columnwidth]{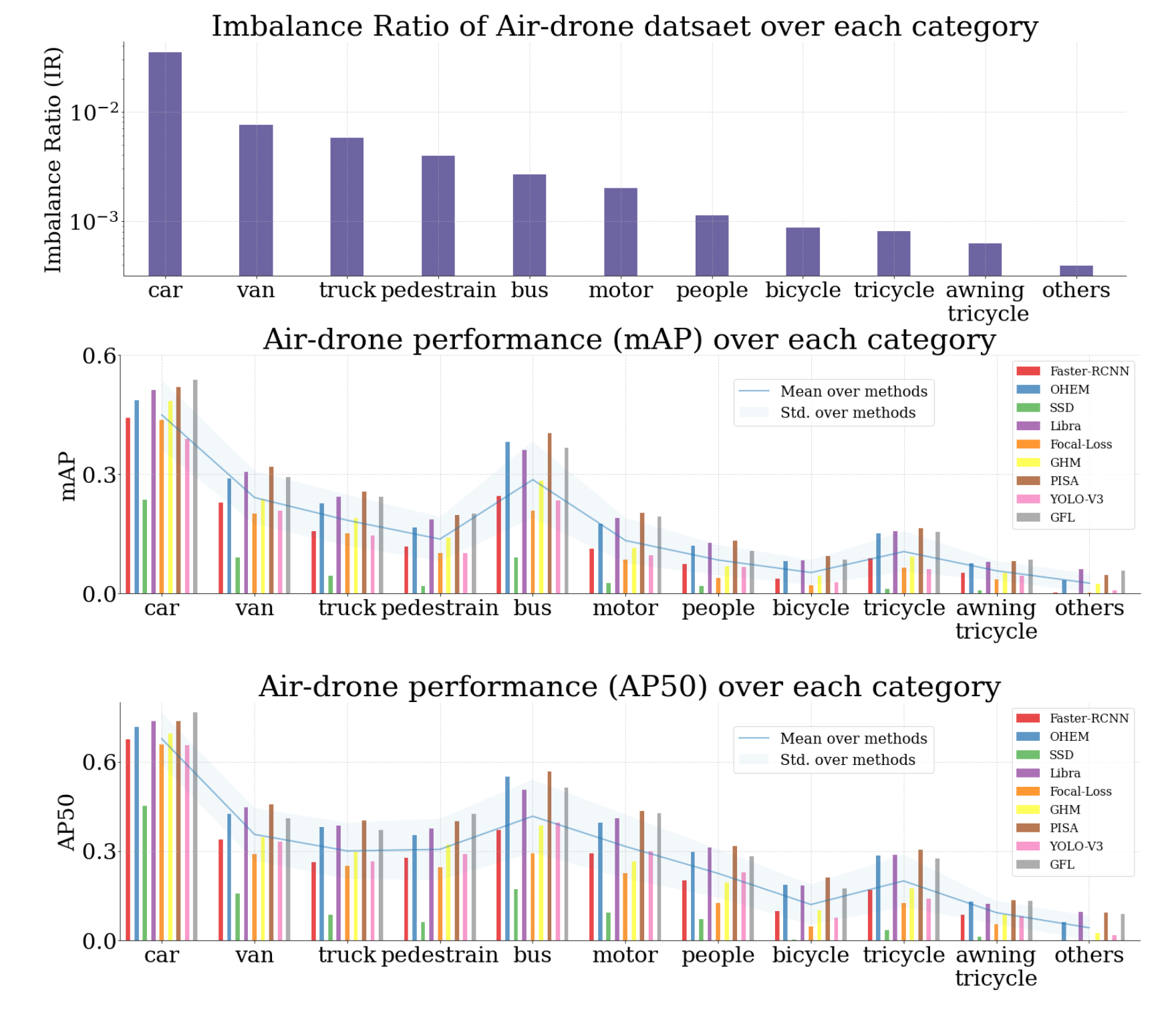}
    \caption{Visualization of the detection results on the Air-drone dataset; the purple bars are the IR for each category of the object, and the curves are the average mAP and AP50 among all the methods for each category.}
    \label{fig:airdrone-results}
\end{figure*}

For the Air-drone dataset, there is a trend that the smaller the imbalance ratio, the worse the detection mAP, as seen in Figure \ref{fig:airdrone-results}. The categories with the lowest imbalance ratios are “awning-tricycle” and ``tricycle'', which have lower mAPs and are hard to detect, while the category ``car'' has the highest mAP along with its lowest IR. On this growing curve, it should be noted that some categories are outliers, such as ``bus'', which has a higher mAP, than the categories with smaller IR as ``truck'' and ``van''. The three best performers on this dataset are Libra-RCNN, PISA, and GFL, with mAP as 0.21, 0.22, and 0.212, respectively.

\section{Conclusions and Discussions}
\subsection{Discussion related to the F-B imbalance.}
Based on the experiments performed in our study, we conclude that:
\begin{enumerate}
  \item	The deterioration of detection performance caused by F-B imbalance is substantial.
\item	If an object type is set, the model’s detection performance decreases with increased imbalance, working either the size or number of the objects changes. A rare exception occurs when there are numerous objects resulting in excessive overlap.
\item	Though typically only the object size is considered in the context of F-B imbalance (the “small object detection” problem), it is only one of the factors for the F-B imbalance that makes the detection task challenging.
\item	The detection performance positively correlates with the object size and density and benefits more when the F-B imbalance issue is severe.
\item	The detection performance is less affected by the F-B imbalance issue when the dataset size grows larger.
\item	With a fixed increasing ratio of IR, it is preferable to increase the size of objects.
\item	The object types and method selection can have a great impact on the detection performance. IR is an important but not exclusive factor for detection performance.
\end{enumerate}
We would explain to them as follows:

For (1) and (2), from the previous experiments, a growing trend of mAP with respect to the increasing IR is observed, as seen in Figure \ref{fig:size-results} and \ref{fig:number-results}. Increasing the object size and adding the object density both contribute to improvements in detecting performance. Although it can be difficult to change the density of the object distribution when working with a custom dataset, it is possible to combat the F-B imbalance by ``editing'' the size of the objects in the background, for example, by cropping images into patches and thus allow the network to pay attention to larger target areas.

For (3), as we previously mentioned, some earlier works concentrated on the F-B imbalance on the ``small object'' issues. The detection performance for small, dense objects (Img-N-Balloon-32-size-32) has a higher performance than datasets with a more serious F-B imbalance but larger object sizes, like (Balloon-1-S-64), as seen in Figure \ref{fig:ir-results}. When a dataset has a high density of small objects, the overall detection performance is satisfying because the F-B imbalance is not severe. Even though there are some additional potential problems brought on by small objects, such as difficulties capturing object pattern features, these problems do not significantly affect detection performance in our experiment settings. Therefore, we claim that small and sparse objects reflect the real challenge in object detection.

With respect to point (4), a larger object size coupled with an increased number of objects results in a greater overall quantity of positive samples derived from the dataset. This, in turn, lessens the severity of the foreground-background (F-B) imbalance. This outcome, in our examination, can be attributed to two principal factors. Firstly, a reduced count of positive samples is generally a consequence of small object sizes and a lesser quantity of objects. Even though a range of sampling methods may be employed to tackle the F-B imbalance, these strategies essentially implement a down-sampling tactic. That is to say, while they decrease the influence of background (negative) samples, they do not rectify the problem when there is a critically low number of positive samples. This meager number of positive samples is insufficient overall to accurately capture the features. In such cases, augmenting the IR, either by enlarging the object sizes or increasing the object quantity, could yield a greater number of positive samples suitable for learning. However, when IR is high, IR is no longer the main problem hindering the detection performance, and the detection performance will be relatively stable at this time and less affected by the F-B imbalance. Secondly, if small object sizes primarily bring on the F-B imbalance, it is more challenging to obtain the object's context information and feature maps accurately. On the other words, enlarging the object can sharply increase its features to be obtained, such as recognizable intricate patterns. However, when the object is big enough, the network can capture enough feature expressions, making it less effective to further the object’s size.

For (5), a similar explanation from the perspective of the number of positive samples can suffice. Regardless of the technique used to address the F-B imbalance, when the dataset size (N) was large, there would be plenty of positive samples and negative samples in total despite the F-B imbalance. No matter which strategy was selected, there are enough positive samples to learn the desired features. This phenomenon also hinted that collecting more data is an effective way to combat the influence of the F-B imbalance. This can improve learning about the characteristics of small or sparse objects and improve detection performance. Simple strategies to alleviate the F-B imbalance, such as random sampling for halving the F-B imbalance, are sufficient to perform well when the dataset size is large, and the choice of methods may be less important.

For (7), when collecting extra data is infeasible, and the dataset size is small, as seen in the results of our experiments in Figure \ref{fig:size-results}, method selection is an important factor for detection performance. The detection performances on various categories of objects also vary significantly even when exposed to a similar level of infrared radiation, according to experimental results from real-world datasets such as COCO Figure \ref{fig:coco-size-results} and air-drone (Figure \ref{fig:airdrone-results}). This is understandable, given that different visualization features of the objects would naturally result in different levels of difficulty. We could say here that IR generally has a positive correlation with the mAP. Still, the severity of the correlation highly depends on the object category and the method selection.

\subsection{Discussion related to the method selection}
Based on the previous experiment results, we found that:
\begin{enumerate}
 \item	When the dataset size is small, the choice of methods significantly influences the detection performance for an F-B imbalanced dataset. When your dataset size is large, the choice of methods is not significantly influential.
 \item	When the dataset size is small, the SSD method is  more susceptible to object size changes, and the Faster-RCNN method is more susceptible to object density changes.
 \item	The ability to address the F-B imbalance issue for Focal-loss and the soft-sampling methods GHM and GFL are relatively unstable on smaller dataset sizes. Thus, repeated experiments to get an optimal model are necessary when choosing these methods for training a custom dataset.
 \item	Among all the selected methods, Libra-RCNN and PISA perform best in handling the F-B imbalance, no matter the size of the datasets or the source of the F-B imbalance.
 \item	One-stage detectors have a smaller model size and better computational efficiency when the model size, model training, and inference time are taken into account. SSD is the model with the smallest model size and the fastest inference speed, and GHM and GFL are relatively good methods with a smaller model size and decent performance on F-B
imbalanced datasets.
 \item	In order to combat an F-B imbalanced dataset, generative methods have great potential to improve detection performance.

\end{enumerate}

For (1), seen in Figure \ref{fig:method-results}, when the dataset size was small, the effectiveness of the detection was significantly impacted by the method that was selected. Taking N-50 as an example, these methods tend to split into two groups, where SSD (one-stage), Faster-RCNN (random sampling), GHM (one-stage), and Focal-loss (one-stage) perform similarly unsatisfactorily, and the mAP was only about half of the other methods. When the dataset size was large enough, such as N=500, the performance of these methods tended to be similar, except for the SSD method, which has a gap compared to the other methods.

For (2) and (3), seen from the results of the balloon datasets, Figure Figure \ref{fig:method-results} (N=50), SSD exhibits a trend of rising and then falling, i.e., detection performance reaches its highest at object size 64. This trend is also seen in the COCO dataset (Figure \ref{fig:method-coco}) for the categories ``cup'' and ``bottle''. We believed this was related to its fixed size as a single shot 1-stage detector, which detects boxes in two fixed sizes. SSD uses 4*4 and 8*8 feature maps and objects with sizes S-128 are larger than the fixed size of the gridded boxes for SSD. As we noticed in Figure \ref{fig:method-results} (N=50), adding object density significantly impacted the Faster-RCNN method when the dataset was small. We assumed that for the ``random sampling'' method as Faster-RCNN, down-sampling the number of negative samples to keep the positives and negatives as 1:3 is more sensitive to the total number of positive samples in the datasets. Increasing the density of objects increases not only the number of incremental samples but also the total number of samples that can be passed to the second stage of the detector, allowing the network to learn a more comprehensive set of features. We noted that Focal-loss, a soft-sampling method for adjusting the weight by an exponential ratio, is unsatisfactory when the dataset size is small (N=50, N=100 in Figure \ref{fig:method-results}), which we believe that the probability-based re-weighting method is fragile \cite{li2020generalized} when the dataset size was small and easily affected by noisy data points, which might make the overall optimization of the model poor. This instability also affects the standard deviation of these methods on repeated experiments, shown in Figure \ref{fig:method-results}) and Table \ref{tab: dbt}, where the best-performed model even has twice the mAP as the worst model. Thus when choosing these methods for a custom dataset, repeated experiments for choosing the optimal model or reporting a fair performance for these methods are necessary.

For (4), we found that the Libra-RCNN and PISA methods achieved the relatively best performance among all the chosen methods, which could be observed in the previous experiments on both synthetic and real-world datasets. So we assumed that although the problem of F-B imbalance is common in real life and there are various strategies dedicated to this problem, the Libra-RCNN and PISA methods are two of the reliable ones. They take into account not only the mitigation of the classification imbalance of foreground and background samples but also other aspects brought by the F-B imbalance, such as the challenges of small feature size and inter-object occlusion with a feature balance. Moreover, both of them achieved good performance when the size of the dataset was very small, e.g., when there were only 50 collectible training data. This is because the IoU-balanced sampling in Libra takes into account the impact of noise labels on the detection performance so that it is not swayed by the outliers even when the dataset size is small. And PISA is also an IoU-hierarchical-based ranking to favor prime samples, which can also make better use of small dataset sizes.

For (5), certainly, we believe that in evaluating a method, its mAP is not the only metric to consider. For instance, SSD, although it has been in a place of unsatisfactory performance on average in our experiments, serves as a concise 1-stage detector with the smallest model size and fast training and inference speeds, as seen in Table \ref{tab: dbt}, making it possible that it may also be chosen when computational resources are limited, and model speed is required. Also, in general, one-stage detectors have a smaller model size and faster inference speed, making GHM and GFL two methods promising to balance computational efficiency and model performance.

For (6), as we learned from Section 5.3, shown in Figure \ref{fig:ir-results}, more objects could significantly enhance the detection performance on severe F-B imbalanced data sets. Although ``editing'' the dataset while training is impossible with conventional methods, there is a lot of potential for future generative methods, which could add “fake” objects to the background and thereby improve the detection performance.

To conclude, in this work, we conducted a comprehensive analysis and many experiments from a synthesis dataset and several real-world datasets. Experiment results demonstrate that the F-B imbalance is detrimental to object detection performance. Our thirteen statements provide easy-to-implement recommendations for addressing the F-B imbalance issue, which can be applied to real-world scenarios.

\section{Acknowledgments}
Research reported in this publication was supported by the National Institute Of Biomedical Imaging And Bioengineering of the National Institutes of Health under Award Number R01EB021360. The content is solely the responsibility of the authors and does not necessarily represent the official views of the National Institutes of Health.

\bibliographystyle{unsrtnat}
\bibliography{mybibfile}  






\end{document}